\documentclass[11pt]{article}
\usepackage[utf8]{inputenc}
\usepackage[T1]{fontenc}
\usepackage{amsmath,amssymb,amsthm}
\usepackage{geometry}
\usepackage{hyperref}
\usepackage{graphicx}
\usepackage{indentfirst,csquotes}
\usepackage{xcolor,paralist,fancyhdr,etoolbox}
\usepackage{tikz}
\usetikzlibrary{positioning, arrows.meta}
\usepackage{lipsum}
\usepackage{float}
\usepackage{subcaption}
\usepackage{titlesec}
\usepackage{secdot}
\usepackage{microtype}
\usepackage{ragged2e}
\usepackage{tabularx}
\usepackage{booktabs}
\usepackage{placeins}
\usepackage{authblk}
\usepackage{orcidlink}
\allowdisplaybreaks

\newtheorem{theorem}{Theorem}

\newcommand{\subsubsubsection}[1]{\paragraph{#1}}
\titleformat{\paragraph}
  {\normalfont\normalsize\bfseries}
  {\theparagraph}
  {1em}
  {}
\titlespacing*{\paragraph}
  {0pt}{3.25ex plus 1ex minus .2ex}{1.5ex plus .2ex}
  
\begin{document}

\title{SW-KAN: Kolmogorov--Arnold Networks with Stieltjes--Wigert $q$-Orthogonal Polynomials}

\author{
	Amirhosein Azarpour\textsuperscript{1, \dag} \orcidlink{0009-0004-1738-1919}, 
	Seyyed Moein Kazemi\textsuperscript{2, \dag} \orcidlink{0009-0001-1860-7236} \\
	\textsuperscript{1} Faculty of Mathematical Sciences, Department of Computer Science, Shahid Beheshti University, Tehran \\
	\textsuperscript{2} Faculty of Mathematical Sciences, Shahid Beheshti University, Tehran \\
	\texttt{\{am.azarpour, seyy.kazemi\}@mail.sbu.ac.ir}
	\thanks{\textsuperscript{\dag} These authors contributed equally to this work.}
}

\maketitle

\footnotetext[1]{The source code and pretrained models are available at: 
\href{https://github.com/amirhoseinazarpour/SW-KAN}{https://github.com/amirhoseinazarpour/SW-KAN}}

\begin{abstract}
Kolmogorov-Arnold Networks (KANs) represent a paradigmatic shift in deep learning by replacing fixed node activations with learnable univariate functions on edges, offering enhanced interpretability and parameter efficiency. While recent polynomial-based KAN variants have addressed the computational overhead of original B-spline implementations, they introduce a fundamental yet underexplored challenge: the domain mismatch between unbounded real-valued inputs and the bounded or semi-infinite support of orthogonal polynomial bases. To address this limitation, we propose the Stieltjes-Wigert Kolmogorov-Arnold Network (SW-KAN), a novel architecture that employs Stieltjes-Wigert $q$-orthogonal polynomials defined on the semi-infinite domain $(0,\infty)$. We introduce a smooth exponential-of-tanh mapping that stably bridges the domain gap while preserving well-conditioned gradients, and leverage a numerically stable three-term recurrence that evaluates polynomial expansions in $\mathcal{O}(N)$ operations without special-function calls. Through comprehensive experiments spanning image classification and continuous function approximation, we demonstrate that SW-KAN consistently achieves superior accuracy-efficiency trade-offs across diverse tasks. Our results show that the log-normal weight structure and learnable $q$-parameter of Stieltjes-Wigert polynomials provide a distinct inductive bias that enables robust performance under resource-constrained conditions, including reduced feature dimensionality and limited training data. The proposed architecture not only outperforms established polynomial KAN baselines on standard benchmarks but also exhibits strong representational capacity for approximating complex multivariate functions with remarkably few parameters. These findings establish SW-KAN as a compelling and versatile alternative for efficient function approximation and classification, with particular promise for real-world applications where computational resources and labeled data are limited.
\end{abstract}

% کلیدواژه‌ها با فرمت ساده (جایگزین IEEEkeywords)

\section{Introduction}
\label{sec:introduction}

Multi-Layer Perceptrons (MLPs) have served as the foundational building blocks
of modern deep learning, with their representational capacity grounded in the
universal approximation theorem~\cite{goodfellow2016deep}.
Despite their broad adoption, MLPs assign fixed activation functions to nodes
and rely on learned scalar weights on edges, a design that can require large
parameter counts to capture intricate nonlinear relationships and offers
limited mathematical transparency.
A notable paradigm shift was introduced by Liu \textit{et al.}~\cite{liu2024kan},
who proposed Kolmogorov-Arnold Networks (KANs) as a principled alternative
to MLPs.
Motivated by the Kolmogorov-Arnold Representation Theorem, which asserts that
any continuous multivariate function $f:[0,1]^n \to \mathbb{R}$ can be expressed
as a finite superposition of continuous univariate functions~\cite{schmidt2021kar},
KANs replace fixed nodal activations with learnable univariate functions placed
on the network's \emph{edges}.
This architectural choice decouples the representational role of activations
from the aggregation role of nodes, enabling more compact models with favorable
accuracy-parameter trade-offs and improved structural interpretability~\cite{liu2024kan}.

Following the introduction of the original KAN, which parametrizes edge
activations with B-spline curves, a rapidly growing body of research has
explored alternative basis functions that preserve the KAN framework's
interpretability while addressing practical limitations of splines.
Liu \textit{et al.}~\cite{liu2024kan2} subsequently extended the framework in
KAN~2.0, introducing multiplication nodes and a symbolic compiler to further
bridge the gap between neural computation and analytical scientific expression.
Seydi~\cite{seydi2024poly} conducted a systematic comparative study of 18
polynomial families as KAN basis functions encompassing orthogonal polynomials,
hypergeometric polynomials, $q$-polynomials, and Fibonacci-related
polynomials and evaluated their performance on the MNIST handwritten digit
classification benchmark.
Concurrently, Sidharth \textit{et al.}~\cite{chebykan2024} proposed the
Chebyshev KAN (ChebyKAN), which leverages the minimax optimality and three-term
recurrence of Chebyshev polynomials of the first kind to parametrize edge
activations, demonstrating improved parameter efficiency over the original
B-spline formulation.
The fractional KAN (fKAN) of Aghaei~\cite{aghaei2025fkan} further extended
this line of inquiry by incorporating trainable fractional-orthogonal Jacobi
basis functions, exploiting their non-polynomial behavior and analytically simple
derivative formulas to improve accuracy across both supervised and
physics-informed tasks.
Aghaei~\cite{aghaei2024rkan} additionally investigated rational basis functions
for KANs via Pad\'{e} approximation and rational Jacobi functions, establishing
the rational KAN (rKAN) and reporting competitive classification accuracy on
standard benchmarks.
Collectively, these works establish a clear trajectory: classical orthogonal
and rational polynomial families are increasingly preferred over B-splines as
the primary parametrization strategy for KAN edge activations.

However, several substantive limitations persist in the current landscape of
polynomial KAN variants.
The original B-spline formulation incurs significant computational overhead,
as maintaining knot vectors and evaluating piecewise polynomial segments
complicates both the forward pass and gradient
computation~\cite{liu2024kan, aghaei2024rkan}.
While classical polynomial KANs attenuate these costs, they frequently exhibit
sub-optimal accuracy-parameter trade-offs: increasing polynomial degree to
improve approximation fidelity inflates the parameter count without proportional
gains in task accuracy, as demonstrated by the degree-accuracy ablation reported
for ChebyKAN~\cite{chebykan2024}.
Furthermore, a structural challenge that receives insufficient attention in the
existing literature is the \emph{domain mismatch problem}.
Classical orthogonal polynomial families such as Chebyshev and Legendre are
defined on the bounded interval $[-1,1]$, while layer inputs in deep networks
are, in general, unbounded real-valued quantities.
Existing approaches commonly address this by applying a $\tanh$ normalization
to compress inputs into $[-1,1]$; however, this compression can saturate in the
tails and introduce gradient attenuation for large-magnitude
activations~\cite{chebykan2024}.
For polynomial families defined on semi-infinite or otherwise non-standard
domains whose weight structures may offer distinct inductive biases no
principled and numerically stable domain-bridging strategy has been
systematically proposed or evaluated within the KAN framework. Recent work on learnable recursive polynomial bases for KANs~\cite{azarpour2026reckan} has shown the promise of adapting the basis-generating rule itself.

To address these limitations, we propose the \textbf{Stieltjes-Wigert KAN
(SW-KAN)}, a novel KAN architecture that replaces B-spline edge activations
with Stieltjes-Wigert (SW) $q$-orthogonal polynomials, which are orthogonal
with respect to a log-normal weight function on the semi-infinite support
$(0,\infty)$, providing a mathematically distinct inductive structure
relative to classical bounded-domain polynomial families. To resolve the
mismatch between unbounded real-valued layer inputs $x \in \mathbb{R}$ and
this strictly positive support, we introduce a smooth, invertible
\emph{exponential-of-tanh} domain mapping $\phi(x) = e^{\tanh(x)}$, which
bijectively maps $\mathbb{R}$ onto the compact subinterval
$(e^{-1}, e) \subset (0,\infty)$ while maintaining bounded, well-conditioned
gradients. The SW polynomial expansion along each edge is then computed via a
three-term recurrence, evaluating a degree-$N$ basis in only $\mathcal{O}(N)$
operations per edge and bypassing costly hypergeometric special-function
evaluations.

The main contributions of this work are:

\begin{itemize}
    \item \textbf{SW-KAN Architecture:} learnable edge activations
    parametrized by Stieltjes-Wigert $q$-orthogonal polynomials, introducing
    a semi-infinite log-normal-support polynomial family into the KAN design
    space.

    \item \textbf{Exponential-of-Tanh Domain Mapping:} a smooth, invertible
    transformation that stably projects unbounded real-valued layer inputs
    into the semi-infinite support $(0,\infty)$ of the SW polynomials,
    preserving well-conditioned gradients and mitigating numerical
    saturation.

    \item \textbf{Numerically Stable Recurrence-Based Evaluation:} an
    efficient, numerically stable recurrence that computes the full
    degree-$N$ SW expansion per edge in $\mathcal{O}(N)$ operations, without
    reliance on hypergeometric function libraries.

    \item \textbf{Empirical Validation Across Diverse Tasks:} on
    MNIST~\cite{seydi2024poly}, SW-KAN achieves a competitive test accuracy
    of $98.24\%$ while outperforming all 18 polynomial KAN baselines
    evaluated therein, including Gottlieb-KAN and Vieta-Pell-KAN, at a
    comparable parameter count. We further validate SW-KAN under
    resource-constrained conditions on Fashion-MNIST (reduced feature
    dimensionality, limited training data) and on continuous multivariate
    function approximation, showing its advantages generalize beyond a
    single benchmark and task type.
\end{itemize}

The remainder of this paper is organized as follows: Section~\ref{sec:related_works} reviews the mathematical foundations of KART, Stieltjes-Wigert $q$-orthogonal polynomials, and related KAN architectures; Section~\ref{sec:methodology} details the SW-KAN architecture, including the domain mapping and recurrence-based forward pass; Section~\ref{sec:experiments_results} presents results on MNIST, Fashion-MNIST, and continuous function approximation against polynomial-KAN and B-spline-KAN baselines; Section~\ref{sec:conclusion} concludes with findings and future directions.

\section{Related Work}
\label{sec:related_works}

\subsection{Theoretical Foundations and the Limits of Multi-Layer Perceptrons}

The theoretical grounding of neural-network function approximation rests on
the Universal Approximation Theorem, which establishes that Multi-Layer
Perceptrons (MLPs) with a single hidden layer of sufficient width can
approximate any continuous function on a compact domain to arbitrary
precision~\cite{hornik1989}. Despite this expressive guarantee, the MLP
architecture fixes its non-linearities at the \emph{nodes}, forcing
representational capacity to be purchased exclusively through increased width
or depth. This design choice imposes well-documented parameter inefficiency:
for a given approximation accuracy, MLPs often require substantially more
trainable parameters than the target function's intrinsic complexity would
suggest, while simultaneously obscuring the mathematical structure of the
learned mapping~\cite{liu2024kan}. A parallel theoretical tradition,
originating with Kolmogorov's resolution of Hilbert's thirteenth problem,
offers a structurally distinct viewpoint: the Kolmogorov-Arnold Representation
Theorem (KART) asserts that any continuous function
$f\colon [0,1]^n\!\to\!\mathbb{R}$ can be expressed as a finite superposition
of continuous univariate functions~\cite{kolmogorov1957}. Montanelli and
Du~\cite{montanelli2020} demonstrated that a constructive proof of KART
directly yields sharp approximation-error bounds for deep ReLU networks,
providing a principled alternative to empirical depth-width heuristics. The
representational analysis was further refined in~\cite{chen2027LegendreKANHighAccuracy}, where it
was shown that a modified form of the Kolmogorov-Arnold decomposition transfers
the smoothness of the target function to its univariate components a property
that is essential for stable numerical optimisation. Together, these results
recast KART from a pure-mathematical theorem into a tractable blueprint for
interpretable network design.

\subsection{Kolmogorov-Arnold Networks: Activations on Edges}

Motivated by KART, Liu~et~al.~\cite{liu2024kan} introduced Kolmogorov-Arnold
Networks (KANs), which relocate learnable non-linearities from the nodes to
the \emph{edges}: every scalar weight in a conventional MLP is replaced by a
trainable univariate B-spline function, so each connection itself performs a
non-linear transformation before the outputs are summed at the receiving node.
This seemingly localised change has two empirically validated consequences.
First, for equivalent approximation accuracy, KANs require substantially fewer
parameters than comparable MLPs on function-regression and partial differential
equation (PDE) benchmarks~\cite{liu2024kan}. Second, the learned edge
functions can be directly inspected and symbolically extracted, yielding models
that are interpretable in a mathematically meaningful sense rather than merely
post-hoc explainable. Liu~et~al. extended the framework in KAN
2.0~\cite{liu2024kan2} by introducing multiplication nodes, a KAN-to-symbolic
compiler, and tree-graph converters that enable the automated discovery of
physical conservation laws and Lagrangian structures. The applied utility of
the edge-activation paradigm was corroborated by Wang~et~al.~\cite{wang2025kinn},
whose Kolmogorov-Arnold Informed Neural Network (KINN) integrated KAN layers
into a physics-informed training loop and demonstrated that KINNs
outperform MLP-based PINNs in accuracy and convergence speed across a range of
solid-mechanics PDEs. Furthermore, Bozorgasl and Chen~\cite{bozorgasl2024wavkan}
explored wavelet-parameterised edge functions, showing that the flexibility of
the KAN edge is not restricted to splines but can accommodate any sufficiently
regular univariate family, thereby generalising the architectural principle
introduced in~\cite{liu2024kan}.

\subsection{Polynomial KAN Variants: Efficiency Gains and Persistent Bottlenecks}

The computational characteristics of B-spline edge functions, however, present
scalability challenges that have motivated a sustained research effort in
polynomial KAN variants. Specifically, evaluating B-splines of degree $k$ over
a grid of $G$ knots requires $\mathcal{O}(kG)$ recursive operations per edge
that resist straightforward parallelisation, as demonstrated in the
MatrixKAN analysis of Atwell~et~al.~\cite{matrixkan2025}. Furthermore,
the grid-extension procedure which redistributes knots to track the empirical
input distribution during training couples the network's parameter structure
to the data statistics in a manner that complicates mini-batch training and
requires careful scheduling~\cite{ukan2024}. Free-Knots~KAN~\cite{freeknots2025}
partially addressed training instability by analytically bounding the number of
active knots, yet the dependency of approximation quality on input-domain
coverage remained a structural constraint.

In response to these limitations, a body of work has pursued globally defined
orthogonal polynomial bases as drop-in replacements for B-splines on KAN edges.
Seydi~\cite{seydi2024poly} conducted a systematic comparative evaluation of
eighteen distinct polynomial families including classical orthogonal
polynomials (Chebyshev, Legendre, Jacobi, Gegenbauer), hypergeometric
polynomials, $q$-polynomials, and combinatorial families as KAN edge-basis
functions, finding that while certain families achieve competitive accuracy on
image classification benchmarks, no single family consistently attains an
optimal parameter-accuracy trade-off across heterogeneous tasks. The Chebyshev
KAN (Cheb-KAN) proposed in~\cite{chebykan2024} demonstrated that Chebyshev
polynomials of the first kind, by virtue of their orthogonality on $[-1,1]$
with respect to the Chebyshev weight and their minimax approximation property,
yield parameter-efficient edge activations with improved function-approximation
fidelity relative to B-spline KANs of comparable size. This orthogonality
advantage was further exploited by Li~et~al.~\cite{chebpikan2024} in a
physics-informed Chebyshev-KAN for fluid-mechanics PDEs, achieving accuracy
improvements at reduced parameter counts. Concurrently, Aghaei~\cite{aghaei2025fkan}
introduced the Fractional KAN (fKAN), which employs trainable
fractional-orthogonal Jacobi functions as edge activations; the fractional
parameterisation confers non-polynomial behaviour and activity for both
positive and negative inputs, resulting in measurable training-speed gains
over integer-order polynomial variants. Rational extensions were explored by
He~et~al.~\cite{aghaei2024rkan}, whose rKAN substitutes Padé-approximant and
rational Jacobi basis functions for B-splines, and by Aghaei and colleagues
in the context of 3D point-cloud processing~\cite{jacobikan2024}, where Jacobi
polynomials with tunable shape parameters $(\alpha, \beta)$ were evaluated across
multiple special cases (Legendre, Chebyshev, Gegenbauer).

Collectively, these polynomial KAN variants address the computational overhead
of B-spline recursion and grid management. However, they share a common
structural challenge that has received insufficient systematic treatment in the
existing literature: classical orthogonal polynomials are orthogonal with
respect to weight functions supported on \emph{bounded} intervals, most
commonly $[-1, 1]$. When such architectures process inputs drawn from the
unbounded real line $\mathbb{R}$, an explicit domain-mapping step is required
to project each input onto the polynomial's natural domain. Commonly employed
strategies---linear rescaling, sigmoid, or $\tanh$ normalisation---may
saturate the polynomial's approximation range for inputs with heavy-tailed or
skewed distributions, or introduce boundary artefacts near $\pm 1$ where
polynomial oscillations concentrate~\cite{seydi2024poly,chebykan2024}. In the
absence of a numerically principled mapping, evaluations near the boundary of
$[-1,1]$ may exhibit gradient vanishing or amplification, degrading both
training convergence and out-of-distribution generalisation. This
domain-mismatch problem, compounded by residual parameter-accuracy trade-offs,
represents the principal unresolved bottleneck in the polynomial-KAN paradigm.

\subsection{q-Orthogonal Polynomials and the Semi-Infinite Domain Opportunity}

A class of polynomial families that sidesteps the domain-mismatch problem
entirely consists of orthogonal polynomials defined on semi-infinite or
fully unbounded domains. Within the classical theory of $q$-orthogonal
polynomials~\cite{koekoek1996}, the Stieltjes-Wigert polynomials occupy a
distinguished position: they are orthogonal with respect to a log-normal weight
function supported on the semi-infinite interval $(0, \infty)$, rendering them
structurally compatible with positive real-valued inputs without any domain
truncation. Moreover, Stieltjes-Wigert polynomials satisfy a three-term
recurrence relation that admits $\mathcal{O}(N)$ evaluation of the first $N$
basis functions without invoking special-function libraries, a computational
profile that compares favourably with B-spline grids of comparable polynomial
degree and with the trigonometric evaluations required by Chebyshev
recurrences for large $N$~\cite{seydi2024poly}. The $q$-deformation parameter
provides an additional mechanism for adjusting the polynomial's concentration
and oscillatory frequency to match the empirical input distribution, a degree
of adaptability absent from fixed classical families. Furthermore, the
underlying $q$-analytic framework connects Stieltjes-Wigert polynomials to the
broader Askey $q$-scheme~\cite{koekoek1996}, providing a well-developed
approximation-theoretic foundation. Despite these properties, the integration
of $q$-orthogonal polynomial families into KAN edge activations, together with
a principled and numerically stable mechanism for mapping arbitrary
real-valued inputs onto $(0, \infty)$, has not been addressed in the
existing literature. The present work closes this gap by proposing SW-KAN,
which replaces B-spline edge activations with Stieltjes-Wigert $q$-orthogonal
basis functions, employs an exponential-of-$\tanh$ transformation for
bijective and numerically stable domain matching from $\mathbb{R}$ to
$(0,\infty)$, and exploits the three-term recurrence for $\mathcal{O}(N)$
basis evaluation simultaneously resolving the grid-extension overhead of
B-spline KANs and the domain-mismatch instability inherent in
bounded-polynomial KANs.

\section{Methodology}
\label{sec:methodology}

\subsection{Kolmogorov--Arnold Theorem}
\label{subsec:kart}
\vspace{0.25cm}
The Kolmogorov--Arnold Representation Theorem (KART) is a fundamental result
in approximation theory that provides the theoretical foundation for
Kolmogorov--Arnold Networks \cite{kolmogorov1957, arnold2009FunctionsThreeVariables}.

\subsubsection{Statement of the Theorem}

In 1957, Kolmogorov proved that any continuous multivariate function on a bounded
domain can be exactly decomposed into a finite composition of continuous univariate
functions and addition, thereby resolving Hilbert's 13th problem in the
affirmative \cite{kolmogorov1957}. The result was subsequently
refined by his student Arnold \cite{arnold2009FunctionsThreeVariables}.

\begin{theorem}[Kolmogorov, 1957; Arnold, 1957]
For any continuous function $f: [0,1]^n \to \mathbb{R}$, there exist continuous
univariate functions $\Phi_q: \mathbb{R} \to \mathbb{R}$ and
$\varphi_{q,p}: [0,1] \to \mathbb{R}$ such that
\begin{equation}\label{eq:kat}
  f(x_1, x_2, \ldots, x_n)
  = \sum_{q=0}^{2n} \Phi_q\!\left(\sum_{p=1}^{n} \varphi_{q,p}(x_p)\right).
\end{equation}
\end{theorem}

Equation~\eqref{eq:kat} shows that any continuous mapping in $n$ variables can
be constructed entirely from univariate functions and summation no
multiplication between input variables is required
\cite{braun2009constructive}. The representation is exact, not an
approximation. The original construction uses a two-layer structure: an inner
layer of $\varphi_{q,p}$ functions and an outer layer of $\Phi_q$ functions,
with widths $(2n+1)$ and $1$, respectively.

\subsubsection{Significance and Limitations}

The essence of the theorem lies in its ability to transform a function of many
variables into a sum of functions, each depending on only one variable. This
decomposition demonstrates that multivariate continuous functions are not
fundamentally more complex than univariate ones, as they can always be
expressed through compositions of the latter
\cite{kolmogorov1957, schmidt2021kar}.

Despite its theoretical elegance, the KART was long considered impractical for
machine learning due to the potential non-smoothness of the inner functions
$\varphi_{q,p}$ and the lack of constructive algorithms
\cite{girosi1989irrelevant, braun2009constructive}. Early attempts to connect
the KART to neural network architectures were explored by
Hecht-Nielsen~\cite{hechtnielsenkolmogorov1987}, who reinterpreted the theorem
as an existence result for mapping neural networks. This interpretation was
later placed on firmer theoretical footing by
K\r{u}rkov\'{a}~\cite{kurkova1991KolmogorovsTheoremRelevant}, who demonstrated that the
representation could be realized within standard neural network frameworks.

\subsubsection{From KART to KAN}

Recently, Liu et al.~\cite{liu2024kan} revived this line of research by
proposing the Kolmogorov--Arnold Network (KAN), a neural architecture that
replaces fixed activation functions at nodes with learnable univariate
functions on edges. In a KAN, each edge carries a learnable univariate
function and each node performs only summation. For a KAN layer mapping
$\mathbf{x}^{(\ell)} \in \mathbb{R}^{n_\ell}$ to
$\mathbf{x}^{(\ell+1)} \in \mathbb{R}^{n_{\ell+1}}$, the output of the
$j$-th neuron is
\begin{equation}\label{eq:kan_layer}
  x_j^{(\ell+1)}
  = \sum_{i=1}^{n_\ell} \varphi_{j,i}^{(\ell)}\!\left(x_i^{(\ell)}\right),
  \quad j = 1, \ldots, n_{\ell+1},
\end{equation}
where each $\varphi_{j,i}^{(\ell)}: \mathbb{R} \to \mathbb{R}$ is a learnable
univariate activation function placed on the edge connecting node $i$ in
layer~$\ell$ to node~$j$ in layer~$\ell+1$ \cite{liu2024kan}.

While the original KAN parametrizes each $\varphi_{j,i}^{(\ell)}$ using
B-splines, the framework is inherently modular: any family of univariate
functions can serve as the basis \cite{liu2024kan, chebykan2024}.
This modularity has prompted a wave of polynomial KAN variants, including
those based on Chebyshev \cite{chebykan2024}, Jacobi
\cite{aghaei2025fkan}, and Legendre \cite{chen2027LegendreKANHighAccuracy} polynomials.
In this work, we replace B-splines with the Stieltjes--Wigert polynomials, a
$q$-orthogonal family with distinctive computational and theoretical
properties (Section~\ref{subsec:sw_polynomials}).

\subsection{Stieltjes--Wigert Polynomials}
\label{subsec:sw_polynomials}
\vspace{0.25cm}

The Stieltjes--Wigert polynomials are a family of $q$-orthogonal polynomials
first studied by Stieltjes~\cite{stieltjes1894recherches} and
Wigert~\cite{wigert1923fonctions}, residing at the lower level of the
$q$-Askey scheme~\cite{koekoek2010hypergeometric}. They are distinguished by
their connection to the log-normal distribution and the remarkable property
that their associated moment problem is
indeterminate~\cite{christiansen2003moment}. Named after the mathematicians
Thomas Jan Stieltjes and Carl Severin Wigert, these polynomials are especially
important in the context of $q$-series and approximation theory due to their
single-parameter simplicity and efficient recurrence
structure~\cite{ismail2005classical}.

\subsubsection{Definition and Equations}

The Stieltjes--Wigert polynomials $S_n(x; q)$ are defined for a parameter
$0 < q < 1$ and $x \in (0, \infty)$.

\subsubsection{Hypergeometric Form}

The polynomials are given in terms of the basic hypergeometric
function~${}_{1}\phi_{1}$ as~\cite{koekoek2010hypergeometric}:

\begin{equation}\label{eq:sw_def}
  \begin{aligned}
  S_n(x; q)
  &= \frac{1}{(q;\,q)_n}\;
    {}_{1}\phi_{1}\!\left(\frac{q^{-n}}{0}\;;\;q,\;-q^{n+1}x\right) = \frac{1}{(q;\,q)_n}
    \sum_{k=0}^{n} \binom{n}{k}_{\!q}\, q^{k^2}\,(-x)^k,
  \end{aligned}
\end{equation}
where $(q;\,q)_n = \prod_{j=1}^{n}(1 - q^j)$ is the $q$-shifted factorial
(Pochhammer symbol), and
$\displaystyle\binom{n}{k}_{\!q} = \frac{(q;\,q)_n}{(q;\,q)_k\,(q;\,q)_{n-k}}$
is the $q$-binomial coefficient.
\vspace{0.5cm}

The parameter $q$ is related to the weight-function parameter $k > 0$ by

\begin{equation}\label{eq:q_param}
  q = \exp\!\left(-\frac{1}{2k^2}\right).
\end{equation}

\subsubsubsection{Explicit Low-Degree Forms}

The first few Stieltjes--Wigert polynomials are:

\begin{align}
  S_0(x;\,q) &= 1, \label{eq:sw_s0}\\[4pt]
  S_1(x;\,q) &= \frac{1}{1-q}\bigl(1 - q^2 x\bigr), \label{eq:sw_s1}\\[4pt]
  S_2(x;\,q) &= \frac{1}{(1-q)(1-q^2)}
    \bigl(1 - (1+q)\,q^2 x + q^5 x^2\bigr), \label{eq:sw_s2}\\[4pt]
  S_3(x;\,q) &= \begin{aligned}[t]
    \frac{1}{(q;\,q)_3}
    \bigl(1 - &(1+q+q^2)\,q^2 x + (1+q+q^2)\,q^5 x^2 - q^{9} x^3\bigr).
    \end{aligned} \label{eq:sw_s3}
\end{align}

\vspace{0.25cm}

\subsubsection{Recurrence Relations}

The Stieltjes--Wigert polynomials satisfy a three-term recurrence relation,
which is central for their efficient computation. The recurrence
reads~\cite{koekoek2010hypergeometric, koekoek1996}:

\begin{equation}\label{eq:sw_recurrence}
  \begin{aligned}
  -q^{2n+1}\,x\, S_n(x;\,q)
  = (1 - q^{n+1})\,S_{n+1}(x;\,q) - (1 + q - q^{n+1})\,S_n(x;\,q) + q\,S_{n-1}(x;\,q),
  \end{aligned}
\end{equation}
\vspace{0.25cm}

with initial conditions $S_{-1}(x;\,q) = 0$ and $S_0(x;\,q) = 1$.

\vspace{0.25cm}

\paragraph{Monic form.}
Writing $S_n(x;\,q) = \frac{(-1)^n\,q^{n^2}}{(q;\,q)_n}\,p_n(x)$, the monic
polynomials $p_n(x)$ satisfy~\cite{koekoek2010hypergeometric}:

\begin{equation}\label{eq:sw_monic}
  x\,p_n(x)
  = p_{n+1}(x)
    + b_n(q)\,p_n(x)
    + a_n^2(q)\,p_{n-1}(x),
\end{equation}

where the recurrence coefficients are

\begin{equation}\label{eq:sw_coeffs}
  b_n(q) = q^{-2n-\frac{3}{2}}\bigl(1 + q - q^{n+1}\bigr),
  \qquad
  a_n^2(q) = q^{-4n}\bigl(1 - q^n\bigr).
\end{equation}

A key computational advantage is that the coefficients $b_n$ and $a_n^2$ involve
only powers of $q$, enabling evaluation of $S_0, S_1, \ldots, S_N$ in
$\mathcal{O}(N)$ arithmetic operations per input sample, without
special-function calls~\cite{koekoek2010hypergeometric}.

\subsubsection{Orthogonality}

The Stieltjes--Wigert polynomials are orthogonal on the semi-infinite interval
$(0, \infty)$. A distinguishing feature is that the associated moment problem
is \emph{indeterminate}~\cite{christiansen2003moment, stieltjes1894recherches},
meaning that multiple distinct weight functions yield the same orthogonality
relation.

\subsubsubsection{Discrete Weight}

With respect to the discrete measure involving the
$q$-Pochhammer theta product~\cite{koekoek2010hypergeometric}:
\begin{equation}\label{eq:sw_orth_discrete}
  \begin{aligned}
  \int_0^{\infty}
    &S_m(x;\,q)\,S_n(x;\,q)\,
    \frac{dx}{(-x,\,-q/x;\,q)_\infty} = \frac{-\ln q \cdot q^{-n}}{(q;\,q)_\infty\,(q;\,q)_n}
  \;\delta_{mn}.
  \end{aligned}
\end{equation}

\subsubsubsection{Log-Normal Weight}

An equivalent orthogonality holds with respect to the continuous log-normal
weight function~\cite{wigert1923fonctions, christiansen2003moment}:
\begin{equation}\label{eq:sw_weight}
  w(x) = \frac{k}{\sqrt{\pi}}\,x^{-1/2}\,
          \exp\!\bigl(-k^2 \ln^2 x\bigr),
  \qquad x > 0,
\end{equation}
where $q = e^{-1/(2k^2)}$. Under this weight:
\begin{equation}\label{eq:sw_orth_continuous}
  \int_0^{\infty} S_m(x;\,q)\,S_n(x;\,q)\,w(x)\,dx
  = h_n\,\delta_{mn},
\end{equation}
where $h_n > 0$ is the normalization constant. The moments of the log-normal
weight are given by~\cite{christiansen2003moment}:
\begin{equation}\label{eq:sw_moments}
  \mu_n = \int_0^{\infty} x^n\,w(x)\,dx = q^{-n(n+1)/2},
  \qquad n = 0, 1, 2, \ldots
\end{equation}

The indeterminacy of the moment problem means that infinitely many distinct
positive measures on $(0,\infty)$ share the same moment sequence
$\{\mu_n\}_{n \geq 0}$~\cite{christiansen2003moment}. This is a consequence
of the failure of Carleman's
condition~\cite{ismail2005classical, szego1939orthogonal}:
\begin{equation}\label{eq:carleman}
  \sum_{n=1}^{\infty} \mu_n^{-1/(2n)} < \infty.
\end{equation}

\subsection{Justification for the Stieltjes--Wigert Basis vs. Classical Alternatives}
\label{subsec:justification-sw}

While classical orthogonal families such as Hermite (defined on $\mathbb{R}$) or Laguerre (defined on $(0, \infty)$) might appear as intuitive alternatives that either bypass or simplify the domain-mapping requirement, the Stieltjes--Wigert family offers three critical mathematical advantages for parameterizing network edges. 

First, the underlying weight structures differ significantly. Hermite polynomials are orthogonal with respect to a rapidly decaying Gaussian weight, and Laguerre polynomials with respect to an exponential weight. In contrast, Stieltjes--Wigert polynomials are orthogonal with respect to a continuous log-normal weight function (see Equation~\ref{eq:sw_weight}). This heavy-tailed weight structure is inherently better suited to capture the skewed, heavy-tailed activation distributions frequently observed in deep neural networks, mitigating the risk of premature saturation for large-magnitude inputs. 

Second, unlike rigid classical families, the Stieltjes--Wigert basis is a $q$-deformed family. The parameter $q \in (0,1)$ acts as an additional, learnable degree of freedom. As demonstrated in Equation~\ref{eq:sw_recurrence}, $q$ dynamically tunes the spacing of roots and the oscillatory frequency of the basis functions. This allows the network to adapt its local representation resolution to the empirical data distribution during gradient descent, a level of geometric adaptability entirely absent in standard Hermite or Laguerre polynomials. 

Finally, the indeterminate nature of the Stieltjes--Wigert moment problem implies that the basis does not strictly commit to a single unique measure. From a learning perspective, this indeterminacy acts as an implicit regularizer, providing a smoother optimization landscape and reducing the risk of overfitting (Runge's phenomenon) which is a known vulnerability when deploying high-degree classical polynomials on unbounded domains.

\subsection{Domain Handling}
\label{sec:sw-domain-handling}
\vspace{0.25cm}

The Stieltjes--Wigert polynomials are naturally defined on the
semi-infinite domain $(0,\infty)$%
~\cite{stieltjes1894recherches,wigert1923fonctions,koekoek2010hypergeometric}.
However, in a neural network setting, layer inputs $\xi$ typically span
the entire real line $\mathbb{R}$, or at least an unbounded subset
thereof. To bridge this mismatch, we employ an invertible mapping
$\phi:\mathbb{R}\to(0,\infty)$ that transforms arbitrary real-valued
inputs into the support of the polynomials before evaluation%
~\cite{shen2011spectral,boyd2001chebyshev}.

Several classes of smooth, invertible mappings from $\mathbb{R}$ (or a
subset) to $(0,\infty)$ have been explored in the spectral methods
literature. For mappings that first target the interval $(-1,1)$ as an
intermediate step, common choices with a scale parameter $\iota>0$
include:
\begin{align}
  \text{logarithmic:}\quad
  \psi_{\log}(\xi;\iota)
  &= 2\tanh\!\left(\frac{\xi}{\iota}\right) - 1,
  \label{eq:map-log}\\[4pt]
  \text{algebraic:}\quad
  \psi_{\mathrm{alg}}(\xi;\iota)
  &= \frac{\xi - \iota}{\xi + \iota},
  \label{eq:map-alg}\\[4pt]
  \text{exponential:}\quad
  \psi_{\exp}(\xi;\iota)
  &= 1 - 2\exp\!\left(-\frac{\xi}{\iota}\right).
  \label{eq:map-exp}
\end{align}
For the infinite domain $\Omega = (-\infty,\infty)$, typical nonlinear
mappings directly to $(-1,1)$ are:
\begin{align}
  \text{logarithmic:}\quad
  \psi_{\log}^{\infty}(\xi;\iota)
  &= \tanh\!\left(\frac{\xi}{\iota}\right),
  \label{eq:map-log-inf}\\[4pt]
  \text{algebraic:}\quad
  \psi_{\mathrm{alg}}^{\infty}(\xi;\iota)
  &= \frac{\xi}{\sqrt{\xi^{2} + \iota^{2}}}.
  \label{eq:map-alg-inf}
\end{align}
A simple way to obtain a mapping into $(0,\infty)$ is then to compose
one of the above transforms with an exponential lift, e.g.
\begin{equation}
  \phi(\xi)
  = \exp\!\bigl(\gamma\,\psi_{\log}^{\infty}(\xi;\iota)\bigr),
  \qquad \gamma>0,
  \label{eq:sw-domain-map}
\end{equation}
which maps $\mathbb{R}$ smoothly and monotonically onto the bounded
interval $(e^{-\gamma}, e^{\gamma})\subset(0,\infty)$.
This choice guarantees strict positivity, bounds the effective domain to
avoid numerical overflow in the recurrence, and provides well-behaved
gradients during training%
~\cite{shen2011spectral}.
In our experimental configuration, we set the domain mapping hyperparameters to $\gamma=2$ and $\iota=1$. While these parameters could theoretically be treated as learnable weights, this specific static configuration is mathematically motivated by the root distribution of the Stieltjes--Wigert polynomials. Setting $\gamma=2$ explicitly bounds the transformed inputs within the compact interval $(e^{-2}, e^{2}) \approx (0.135, 7.389)$. This interval is critical because it strategically encapsulates the primary region of oscillation (i.e., the location of the roots) for low-to-moderate degree Stieltjes--Wigert polynomials. By keeping the inputs strictly bounded away from zero and preventing them from reaching extreme positive values, we prevent the polynomial evaluations from entering asymptotic tails where gradients typically vanish or explode. Furthermore, setting $\iota=1$ perfectly aligns the active linear regime of the $\tanh$ function with standard normalized layer inputs, ensuring well-conditioned gradient flow and maximum expressivity during backpropagation without the need for additional grid-extension overheads.

\begin{figure}[H]
    \centering
    
    % --- ردیف اول: ۳ شکل ---
    \begin{subfigure}[b]{0.32\textwidth}
        \centering
        \includegraphics[width=\textwidth]{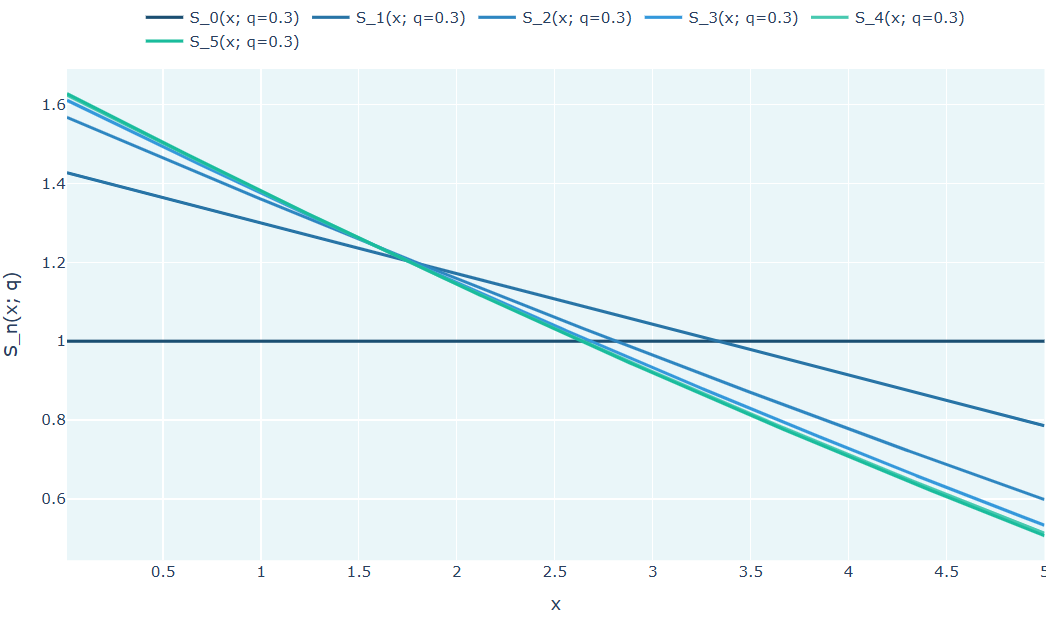}
        \caption{$q=0.3$}
        \label{fig:q03}
    \end{subfigure}
    \hfill
    \begin{subfigure}[b]{0.32\textwidth}
        \centering
        \includegraphics[width=\textwidth]{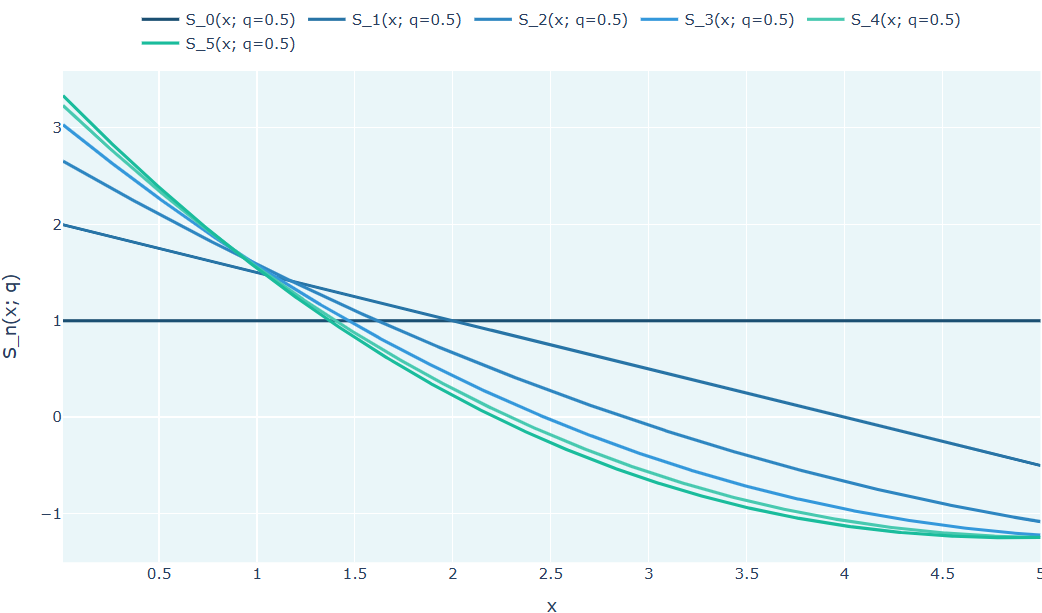}
        \caption{$q=0.5$}
        \label{fig:q05}
    \end{subfigure}
    \hfill
    \begin{subfigure}[b]{0.32\textwidth}
        \centering
        \includegraphics[width=\textwidth]{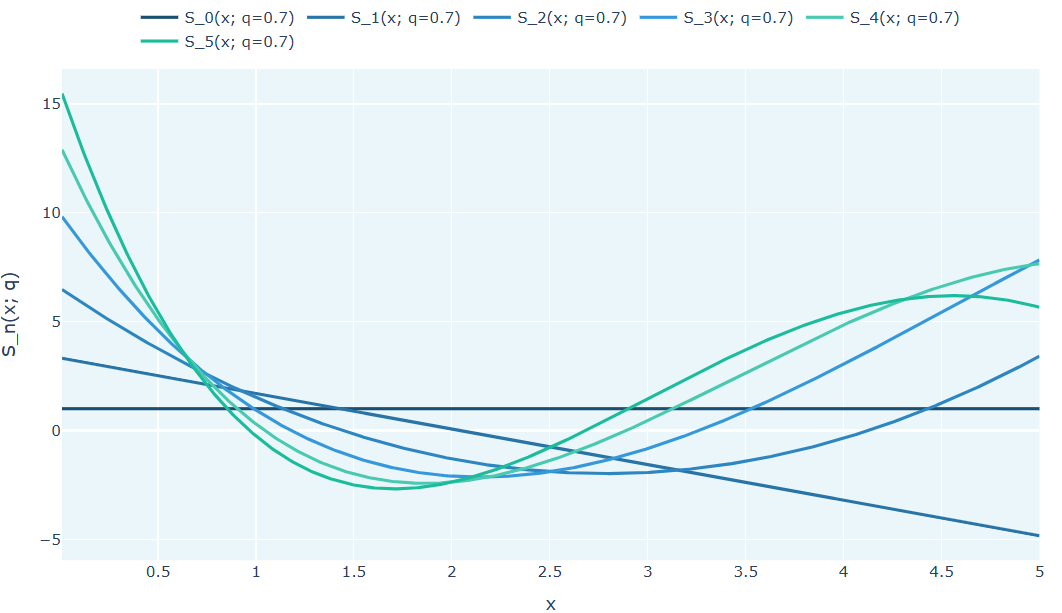}
        \caption{$q=0.7$}
        \label{fig:q07}
    \end{subfigure}
    
    \vspace{0.5cm} % فاصله عمودی بین ردیف بالا و پایین
    
    % --- ردیف دوم: ۲ شکل ---
    \begin{subfigure}[b]{0.32\textwidth}
        \centering
        \includegraphics[width=\textwidth]{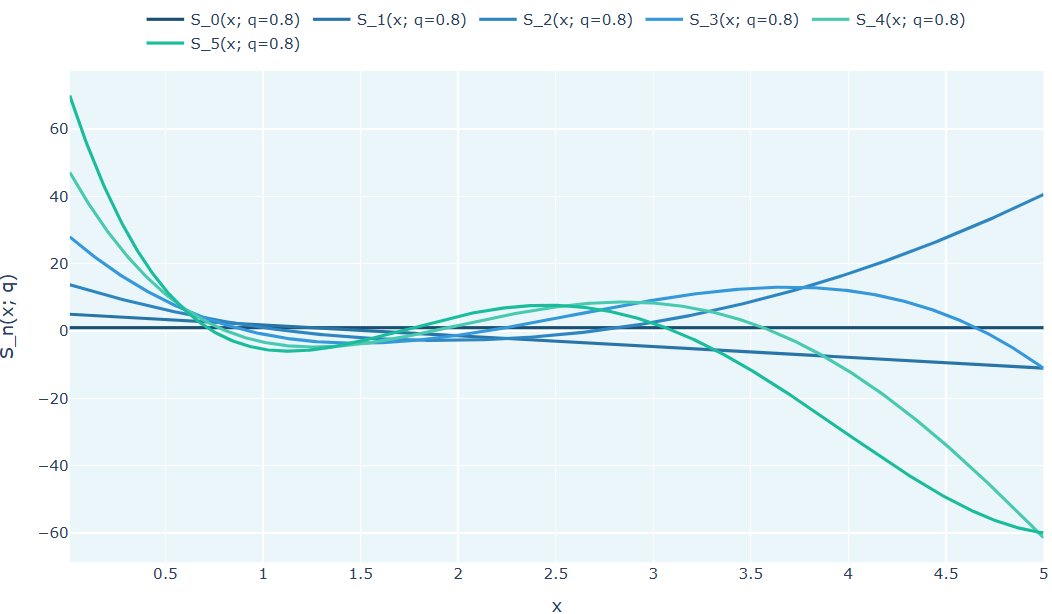}
        \caption{$q=0.8$}
        \label{fig:q08}
    \end{subfigure}
    \hspace{0.5cm} % فاصله افقی بین دو شکل در ردیف پایین
    \begin{subfigure}[b]{0.32\textwidth}
        \centering
        \includegraphics[width=\textwidth]{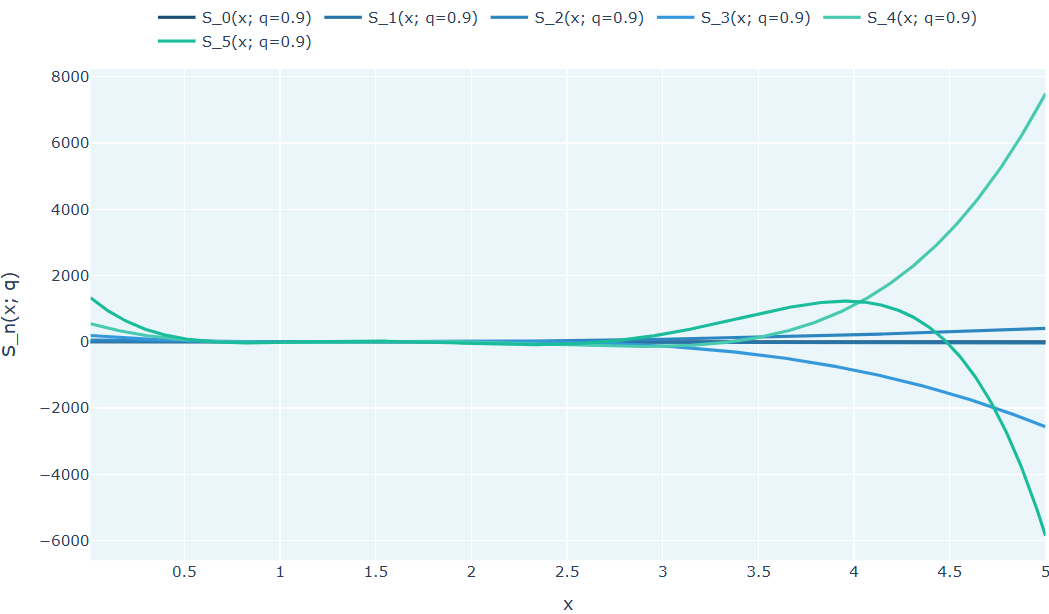}
        \caption{$q=0.9$}
        \label{fig:q09}
    \end{subfigure}
    
    \caption{Stieltjes--Wigert polynomials $S_0$--$S_5$ for different values of $q$.}
    \label{fig:sw_polynomials_all}
\end{figure}

You can see Stieltjes-Wigert polynomials with different values of the parameter $q$ in Figure \ref{fig:sw_polynomials_all}.

\subsection{Stieltjes--Wigert KAN}
\label{sec:sw-kan}
\vspace{0.25cm}

Kolmogorov--Arnold Networks (KANs) replace fixed activation functions on
nodes with learnable univariate functions on edges, yielding compact and
interpretable models for function approximation, symbolic regression,
and scientific computing tasks~\cite{liu2024kan,chebykan2024,aghaei2025fkan,chen2027LegendreKANHighAccuracy}.
In this section we introduce the Stieltjes--Wigert KAN (SW--KAN), a KAN
variant whose edge activations are expanded in the Stieltjes--Wigert
polynomials discussed in Section~\ref{subsec:sw_polynomials}%
~\cite{koekoek2010hypergeometric,ismail2005classical,christiansen2003moment}.

\subsubsection{Edge Activation via Stieltjes--Wigert Polynomials}
\label{subsec:sw-edge-activation}

Consider a KAN layer mapping
\(\mathbf{x}^{(\ell)} \in \mathbb{R}^{n_\ell}\) to
\(\mathbf{x}^{(\ell+1)} \in \mathbb{R}^{n_{\ell+1}}\).
For each edge \((\ell,i) \to (\ell+1,j)\), we parameterize the learnable
univariate activation function using the Stieltjes--Wigert basis as
\vspace{0.25cm}

\begin{equation}\label{eq:sw-edge-activation}
  \begin{aligned}
  \varphi_{j,i}^{(\ell)}(x)
  &=
  \sum_{n=0}^{N}
  c_{j,i,n}^{(\ell)}\,
  S_n\!\bigl(\phi(x);\,q^{(\ell)}\bigr),\qquad i = 1,\dots,n_\ell,\;
  j = 1,\dots,n_{\ell+1},
  \end{aligned}
\end{equation}
where:
\vspace{0.25cm}

\begin{itemize}
  \item \(S_n(\cdot;q)\) are the Stieltjes--Wigert polynomials of
        degree \(n\) with parameter \(0<q<1\)
        (Section~\ref{subsec:sw_polynomials})%
        ~\cite{stieltjes1894recherches,wigert1923fonctions,koekoek2010hypergeometric};
  \item \(\phi:\mathbb{R}\to(0,\infty)\) is a smooth mapping that matches
        the natural domain of the Stieltjes--Wigert polynomials
        (Section~\ref{sec:sw-domain-handling})%
        ~\cite{shen2011spectral,boyd2001chebyshev};
  \item \(c_{j,i,n}^{(\ell)} \in \mathbb{R}\) are learnable expansion
        coefficients, one set per edge \((j,i)\) and per degree \(n\);
  \item \(q^{(\ell)} \in (0,1)\) is a single learnable parameter
        \emph{shared across all edges of layer \(\ell\)}, controlling the
        shape of the basis functions used by that entire layer.
\end{itemize}
A convenient choice of input mapping is the exponential-of-tanh
transform~\cite{shen2011spectral}:
\begin{equation}
  \phi(\xi)
  =
  \exp\!\bigl(\gamma \tanh(\xi / \iota)\bigr),
  \qquad
  \gamma>0,\; \iota>0,
  \label{eq:sw-phi}
\end{equation}
which maps \(\mathbb{R}\) smoothly and monotonically onto
\((e^{-\gamma}, e^{\gamma}) \subset (0,\infty)\), ensuring strict
positivity, boundedness, and stable gradients during backpropagation.

\subsubsection{Recurrence-Based Evaluation}
\label{subsec:sw-recurrence-eval}

Directly evaluating the explicit sum in the hypergeometric definition of
\(S_n(x;q)\) for all \(n=0,\dots,N\) at each edge would be
computationally expensive. Instead, we exploit the three-term recurrence
relation for the monic Stieltjes--Wigert polynomials from
Koekoek--Lesky--Swarttouw~\cite{koekoek2010hypergeometric,koekoek1996}:
\begin{equation}\label{eq:sw-monic-recurrence}
  \begin{aligned}
  x\,S_n(x;q)
  &=
  S_{n+1}(x;q)
  +
  b_n(q)\,S_n(x;q)
  +
  a_n^2(q)\,S_{n-1}(x;q), \\
  &\qquad n \ge 0,
  \end{aligned}
\end{equation}
with initial conditions \(S_{-1}(x;q)=0\) and \(S_0(x;q)=1\), and
recurrence coefficients
\begin{equation}
  b_n(q)
  =
  q^{-2n-3}\,\frac{1+q-q^{n+1}}{2},
  \qquad
  a_n^2(q)
  =
  q^{-4n}\,(1-q^n).
  \label{eq:sw-recurrence-coeffs}
\end{equation}
Equivalently,
\begin{equation}
  S_{n+1}(x;q)
  =
  x\,S_n(x;q)
  - b_n(q)\,S_n(x;q)
  - a_n^2(q)\,S_{n-1}(x;q).
  \label{eq:sw-forward-recurrence}
\end{equation}

For each edge and each input value \(z = \phi(x)\), we compute
\(S_0(z;q),\dots,S_N(z;q)\) by iterating
\eqref{eq:sw-forward-recurrence}, which requires only
\(\mathcal{O}(N)\) arithmetic operations and powers of \(q\), without
special-function calls. This makes SW--KAN significantly cheaper than
KAN variants based on heavier hypergeometric families such as
Askey--Wilson polynomials~\cite{koekoek2010hypergeometric,seydi2024poly}.

\subsubsection{Numerically Stable Evaluation via an Orthonormal Recurrence}
\label{subsec:sw-stable-recurrence}

While \eqref{eq:sw-forward-recurrence} is algebraically exact, iterating
the \emph{monic} recurrence directly in finite-precision arithmetic is
numerically unusable beyond a modest degree. The coefficient
\(a_n^2(q) = q^{-4n}(1-q^n)\) in \eqref{eq:sw-recurrence-coeffs} grows as
\(q^{-4n}\), so for any \(q\) not extremely close to \(1\) the monic
polynomials \(S_n(z;q)\) inflate by many orders of magnitude within a
handful of recursion steps (e.g., \(q=0.35\) and \(n=6\) already yields
\(q^{-4n} \approx 3\times 10^{10}\)). Because \(a_n^2(q)\) enters the
recurrence multiplicatively, this growth compounds across degrees and
routinely produces \texttt{inf}/\texttt{NaN} activations during training
well before the target polynomial degree \(N\) is reached, regardless of
how the input is scaled by \(\phi\) in~\eqref{eq:sw-phi}.

We resolve this by evaluating the mathematically equivalent
\emph{orthonormal} form of the same three-term recurrence rather than the
monic form. Writing \(a_n(q) = \sqrt{a_n^2(q)}\) with \(a_0(q) := 0\), we
define \(T_n(z;q) := S_n(z;q)/\sqrt{h_n}\), where \(h_n>0\) is the norm
constant implied by the orthogonality relation in~\eqref{eq:sw_orth_continuous}.
The sequence \(\{T_n\}\) satisfies
\begin{equation}
  T_{n+1}(z;q)
  =
  \frac{
    \bigl(z - b_n(q)\bigr)\,T_n(z;q)
    -
    a_n(q)\,T_{n-1}(z;q)
  }{
    a_{n+1}(q)
  },
  \label{eq:sw-orthonormal-recurrence}
\end{equation}
with initial conditions \(T_{-1}(z;q) = 0\) and \(T_0(z;q) = 1\), using
the same coefficients \(b_n(q)\) and \(a_n(q)\) as
in~\eqref{eq:sw-recurrence-coeffs}. This is the standard
normalization used to generate orthogonal polynomials stably from Jacobi
recurrence coefficients, underlying, e.g., the node/weight computation in
Gaussian quadrature~\cite{ismail2005classical,koekoek2010hypergeometric}. Rather than multiplying
by the unbounded \(a_n^2(q)\) at each step, \eqref{eq:sw-orthonormal-recurrence}
\emph{divides} by \(a_{n+1}(q)\), which keeps \(T_n(z;q)\) of order
\(\mathcal{O}(1)\) across the working domain \((e^{-\gamma},e^{\gamma})\)
for the full range of admissible \(q\) and degree \(N\), analogous to how
Chebyshev polynomials remain bounded in \([-1,1]\).

Crucially, replacing \(S_n\) with \(T_n\) in the edge activation
\eqref{eq:sw-edge-activation} does not alter the expressiveness of the
model: \(T_n\) differs from \(S_n\) only by the fixed positive scalar
\(1/\sqrt{h_n}\), so \(\{T_n\}_{n=0}^{N}\) spans exactly the same
degree-\(N\) polynomial subspace as \(\{S_n\}_{n=0}^{N}\). Since the
expansion coefficients \(c_{j,i,n}^{(\ell)}\) are learned freely, this
rescaling is absorbed automatically during training and the edge
activation retains the form
\begin{equation}
  \varphi_{j,i}^{(\ell)}(x)
  =
  \sum_{n=0}^{N}
  c_{j,i,n}^{(\ell)}\,
  T_n\!\bigl(\phi(x);\,q^{(\ell)}\bigr).
  \label{eq:sw-edge-activation-stable}
\end{equation}
In practice we additionally add a small \(\varepsilon = 10^{-8}\) under
the square root, \(a_n(q) = \sqrt{a_n^2(q) + \varepsilon}\), to avoid
division by zero when \(q\) is close to \(1\) and \(a_n^2(q) \to 0\). All
SW--KAN results reported in Section~\ref{sec:experiments_results} use
\eqref{eq:sw-orthonormal-recurrence} rather than the raw monic
recurrence \eqref{eq:sw-forward-recurrence}; the latter is presented in
Section~\ref{subsec:sw-recurrence-eval} for consistency with the
classical definition of the Stieltjes--Wigert polynomials, but is not
what is executed at training or inference time.

\subsubsection{Learnable Parameters}
\label{subsec:sw-learnable-params}

For a single SW--KAN layer with \(n_\ell\) input nodes and
\(n_{\ell+1}\) output nodes, and maximum polynomial degree \(N\), the
total number of learnable parameters is
\begin{equation}
  n_\ell n_{\ell+1} (N+1)
  \;\; \underbrace{+}_{\text{layer-wise basis parameter}}\;\;
  1
  \;=\;
  n_\ell n_{\ell+1} (N+1) + 1,
  \label{eq:sw-param-count}
\end{equation}
where:
\begin{itemize}
  \item \(n_\ell n_{\ell+1}(N+1)\) parameters correspond to the
        expansion coefficients \(c_{j,i,n}^{(\ell)}\), i.e.\ \(N+1\)
        coefficients for each of the \(n_\ell n_{\ell+1}\) edges;
  \item the additional single parameter \(q^{(\ell)}\) is
        \emph{shared across all edges of the layer} rather than defined
        per edge, controlling the shape of the Stieltjes--Wigert basis
        functions for that layer as a whole.
\end{itemize}
To enforce the constraint \(q^{(\ell)} \in (0,1)\) during
training, \(q^{(\ell)}\) is obtained from a single unconstrained
parameter \(\hat{q}^{(\ell)} \in \mathbb{R}\) per layer via a sigmoid:
\begin{equation}
  q^{(\ell)} = \sigma\!\bigl(\hat{q}^{(\ell)}\bigr),
  \qquad
  \sigma(t) = \frac{1}{1 + e^{-t}}.
  \label{eq:sw-q-sigmoid}
\end{equation}
This simple parameterization keeps \(q^{(\ell)}\) in the admissible
range and allows gradient-based optimization to adjust it jointly with
the coefficients \(c_{j,i,n}^{(\ell)}\). Sharing a single \(q^{(\ell)}\)
per layer, rather than instantiating one per edge, keeps the parameter
overhead of the \(q\)-deformation negligible (one extra scalar per
layer) while still allowing each layer to adapt the concentration and
oscillatory frequency of its Stieltjes--Wigert basis to the empirical
distribution of its inputs.

\subsubsection{Network Computation}
\label{subsec:sw-network-computation}

Given an input vector \(\mathbf{x}^{(\ell)} \in \mathbb{R}^{n_\ell}\), a
SW--KAN layer first maps each component \(x_i^{(\ell)}\) to the
semi-infinite domain via \(\phi\) in \eqref{eq:sw-phi}, then evaluates
the Stieltjes--Wigert expansion on every edge. The output of neuron
\(j\) in layer \(\ell+1\) is
\begin{equation}\label{eq:sw-layer-output}
  \begin{aligned}
  x_j^{(\ell+1)}
  &=
  \sum_{i=1}^{n_\ell}
  \varphi_{j,i}^{(\ell)}\!\bigl(x_i^{(\ell)}\bigr) =
  \sum_{i=1}^{n_\ell}
  \sum_{n=0}^{N}
  c_{j,i,n}^{(\ell)}\,
  S_n\!\bigl(\phi(x_i^{(\ell)});\,q^{(\ell)}\bigr), \\
  &\qquad j = 1,\dots,n_{\ell+1},
  \end{aligned}
\end{equation}
which is a direct specialization of the general KAN layer formula with
a polynomial basis~\cite{liu2024kan,chebykan2024,seydi2024poly}.

Stacking \(L\) such layers yields the full SW--KAN:
\begin{equation}
  \mathrm{SW\text{-}KAN}(\mathbf{x})
  =
  \Phi^{(L-1)} \circ \Phi^{(L-2)} \circ \cdots \circ \Phi^{(0)}(\mathbf{x}),
  \label{eq:sw-kan-network}
\end{equation}
where each \(\Phi^{(\ell)}\) is the function-valued matrix
\(\bigl(\varphi_{j,i}^{(\ell)}\bigr)_{j,i}\) parameterized via
Stieltjes--Wigert polynomials as in
\eqref{eq:sw-edge-activation}. This architecture can be trained using
standard stochastic gradient methods, with the recurrence-based
evaluation ensuring that the additional expressivity of the
Stieltjes--Wigert basis does not incur prohibitive computational
cost~\cite{koekoek2010hypergeometric,liu2024kan}.
\vspace{0.25cm}

\subsection{Theoretical Computational Complexity and Polynomial Degree}
\label{subsec:complexity_and_degree}

To rigorously establish the computational efficiency of SW--KAN, we must quantify the cost of the three-term recurrence relation compared to the original B-spline formulation. Evaluating a B-spline of degree $k$ over a grid of $G$ knots via the Cox--de Boor recursion requires $\mathcal{O}(kG)$ arithmetic operations per edge. Moreover, B-splines necessitate dynamic grid-extension algorithms during training to ensure the knots dynamically cover the shifting empirical activation distributions. This injects significant non-differentiable overhead and memory fragmentation into the training pipeline.

In stark contrast, the Stieltjes--Wigert forward pass relies entirely on the recurrence relation in Equation~\eqref{eq:sw-forward-recurrence} (evaluated in its numerically stable orthonormal form, Equation~\eqref{eq:sw-orthonormal-recurrence}). Since the coefficients $b_n(q)$ and $a_n(q)$ depend solely on the parameter $q$ and can be updated in $\mathcal{O}(1)$ time, evaluating the basis up to degree $N$ strictly requires $\mathcal{O}(N)$ multiplications and additions per edge. This yields a deterministic $\mathcal{O}(N)$ time complexity that trivially parallelizes on modern GPU architectures, completely bypassing knot management and grid updates.

For our medium SW--KAN configuration yielding $105{,}856$ parameters, the maximum polynomial degree is constrained to $N=3$. From an approximation-theoretic standpoint, $N=3$ provides sufficient representational capacity (i.e., cubic non-linearity with up to two inflection points) to capture complex, non-monotonic feature interactions. Simultaneously, restricting the basis to a low degree explicitly prevents the high-frequency oscillatory artifacts (Runge's phenomenon) that typically degrade the generalization of higher-degree polynomial networks, thereby explaining the robust accuracy achieved without requiring massive parameter counts.

\section{Experiments and Results}
\label{sec:experiments_results}
In this section, we evaluate the proposed Stieltjes--Wigert Kolmogorov--Arnold Network (SW--KAN) on the MNIST handwritten digit classification benchmark and compare it against a broad family of polynomial-based KAN models. Our goal is to assess whether replacing B-spline edge activations with Stieltjes--Wigert $q$-orthogonal polynomials yields a better trade-off between accuracy, model size, and training time than existing polynomial KAN variants. We further investigate SW-KAN's performance on the more challenging Fashion-MNIST dataset and its capability in approximating continuous multivariate functions, demonstrating its generalization across diverse tasks.

\subsection{Digit Classification on MNIST}
The MNIST dataset consists of $60{,}000$ training and $10{,}000$ test grayscale images of handwritten digits $0$--$9$, each of size $28\times 28$. We use the standard train/test split without data augmentation. Input images are flattened to $784$-dimensional vectors and linearly normalized to $[0,1]$. The task is to predict the correct digit label for each image.

\subsubsection{Experimental Setup}
We adopt a fully connected KAN backbone with three hidden KAN layers followed by a linear classification layer, closely following the architecture used in prior work on polynomial-based KANs \cite{liu2024kan,seydi2024poly}. Each hidden layer is implemented as a KAN layer in which every edge $(\ell,i)\to(\ell+1,j)$ carries a learnable univariate activation function. In the SW--KAN model, this activation is parameterized by a Stieltjes--Wigert expansion
\[
\phi^{(\ell)}_{j,i}(x)
=
\sum_{n=0}^{N}
c^{(\ell)}_{j,i,n}\,
S_n\!\big(\varphi(x);q^{(\ell)}\big),
\]
where $S_n(\cdot;q)$ are Stieltjes--Wigert polynomials, $\varphi:\mathbb{R}\to(0,\infty)$ is the exponential-of-tanh mapping from Section~\ref{sec:sw-kan}, $c^{(\ell)}_{j,i,n}$ are trainable coefficients, and $q^{(\ell)}\in(0,1)$ is a single trainable shape parameter shared across all edges of layer $\ell$, enforced via a sigmoid reparameterization.

To study the effect of model size, we consider two configurations:
\begin{itemize}
  \item \textbf{SW--KAN (medium)} with $105{,}856$ trainable parameters, chosen to be comparable to the majority of polynomial KAN baselines.
  \item \textbf{SW--KAN (large)} with $219{,}907$ parameters, matching the parameter count of the widest baseline (Gottlieb--KAN).
\end{itemize}
Both configurations use the same depth and layer widths; the parameter budget is controlled by the maximal polynomial degree $N$ and the number of channels per edge.

All models are trained with the Adam optimizer and cross-entropy loss, using a batch size of $128$, an initial learning rate of $10^{-3}$, and a maximum of $50$ epochs. We track the test accuracy at the end of each epoch and report the best test accuracy achieved during training. Training is performed on a single GPU in a PyTorch implementation.

\subsubsection{Baselines}
As baselines, we include the family of polynomial Kolmogorov--Arnold Networks introduced by Seydi \cite{seydi2024poly}, where the edge activations are expanded in a variety of classical and combinatorial polynomial bases. Specifically, we consider Fermat--KAN, AlSalam--Carlitz--KAN, Bannai--Ito--KAN, Boas--Buck--KAN, Boubaker--KAN, Charlier--KAN, Gottlieb--KAN, Heptanacci--KAN, Hexanacci--KAN, Meixner--Pollaczek--KAN, Narayana--KAN, Octanacci--KAN, Pado--KAN, Pentanacci--KAN, Tetranacci--KAN, Tribo--KAN, Vieta--Pell--KAN, and Askey--Wilson--KAN. All of these models are configured to use a similar KAN backbone, differing only in the choice of polynomial basis on each edge and minor adjustments in polynomial degree to keep the parameter count close to $10^5$ (except Gottlieb--KAN). Their MNIST test accuracies and parameter counts are summarized in Table~\ref{tab:swkan-poly-acc-time}.

\begin{figure}[H]
\centering
\begin{tikzpicture}[
    node distance = 8mm and 4mm,
    every node/.style = {font=\tiny\bfseries, align=center},
    input/.style = {draw, rectangle, rounded corners=1.5mm, minimum height=5mm,
                     minimum width=18mm, thick, fill=blue!12, draw=blue!60!black},
    sw/.style = {draw, rectangle, rounded corners=1.5mm, minimum height=5mm,
                  minimum width=14mm, thick, fill=orange!15, draw=orange!60!black},
    norm/.style = {draw, rectangle, rounded corners=1.5mm, minimum height=5mm,
                    minimum width=18mm, thick, fill=green!12, draw=green!60!black},
    output/.style = {draw, rectangle, rounded corners=1.5mm, minimum height=5mm,
                      minimum width=18mm, thick, fill=red!12, draw=red!60!black},
    arr/.style = {-{Stealth[length=1.5mm, width=1.5mm]}, thick, draw=black!50}
]

    % Input Layer
    \node[input] (input) {Input\\ (28$\times$28)};
    
    % SW0, SW1, SW2 (Layer 1)
    \node[sw, below left=of input, xshift=-10mm] (sw0_1) {SW$_0$\\ (L1)};
    \node[sw, below=of input] (sw1_1) {SW$_1$\\ (L1)};
    \node[sw, below right=of input, xshift=10mm] (sw2_1) {SW$_2$\\ (L1)};
    
    \draw[arr] (input.south) -- (sw0_1.north);
    \draw[arr] (input.south) -- (sw1_1.north);
    \draw[arr] (input.south) -- (sw2_1.north);
    
    % LayerNorm1
    \node[norm, below=of sw1_1, yshift=-3mm] (ln1) {LN1\\ (64)};
    
    \draw[arr] (sw0_1.south) -- (ln1.north);
    \draw[arr] (sw1_1.south) -- (ln1.north);
    \draw[arr] (sw2_1.south) -- (ln1.north);
    
    % SW0, SW1, SW2 (Layer 2)
    \node[sw, below left=of ln1, xshift=-10mm] (sw0_2) {SW$_0$\\ (L2)};
    \node[sw, below=of ln1] (sw1_2) {SW$_1$\\ (L2)};
    \node[sw, below right=of ln1, xshift=10mm] (sw2_2) {SW$_2$\\ (L2)};
    
    \draw[arr] (ln1.south) -- (sw0_2.north);
    \draw[arr] (ln1.south) -- (sw1_2.north);
    \draw[arr] (ln1.south) -- (sw2_2.north);
    
    % LayerNorm2
    \node[norm, below=of sw1_2, yshift=-3mm] (ln2) {LN2\\ (10)};
    
    \draw[arr] (sw0_2.south) -- (ln2.north);
    \draw[arr] (sw1_2.south) -- (ln2.north);
    \draw[arr] (sw2_2.south) -- (ln2.north);
    
    % Output Layer
    \node[output, below=of ln2] (output) {Output\\ (10)};
    \draw[arr] (ln2.south) -- (output.north);

\end{tikzpicture}
\caption{SW-KAN architecture for MNIST.}
\label{fig:diagram}
\end{figure}
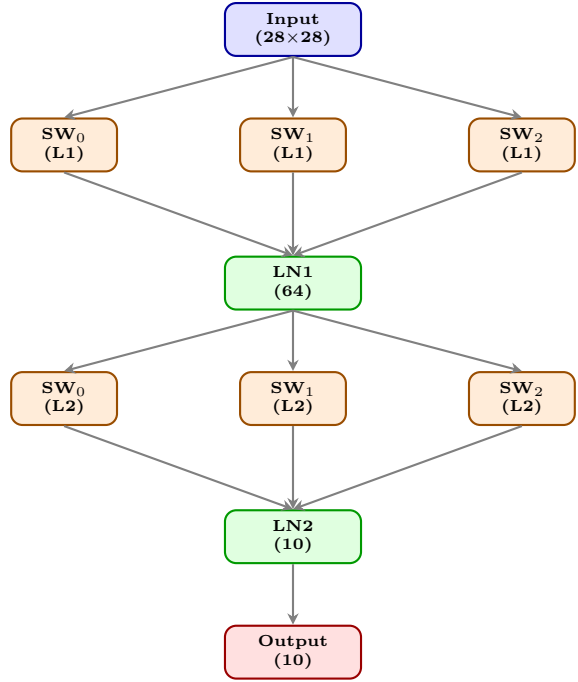

\begin{table}[H]
\centering
\caption{Test accuracy, number of parameters, and total training time (until best epoch) for polynomial-based KAN models and SW-KAN on MNIST. (SW-KAN results are reported at the epoch of best validation accuracy, achieved within 50 epochs.)}
\label{tab:swkan-poly-acc-time}
\begin{tabular}{lccc}
\hline
Model                & Accuracy     & \#Params   & Time (s) \\
\hline
Fermat-KAN           & 0.9619       & 105{,}856  & 757.21   \\
AlSalam-Carlitz-KAN  & 0.9675       & 105{,}862  & 927.30   \\
BannaiIto-KAN        & 0.9670       & 105{,}865  & 1{,}261.58 \\
Boas-Buck-KAN        & 0.9663       & 105{,}880  & 1{,}220.49 \\
Boubaker-KAN         & 0.9731       & 105{,}856  & 807.36   \\
Charlier-KAN         & 0.9726       & 105{,}856  & 853.90   \\
Gottlieb-KAN         & 0.9759       & 219{,}907  & 923.96   \\
Heptanacci-KAN       & 0.9426       & 105{,}856  & 729.22   \\
Hexanacci-KAN        & 0.9641       & 105{,}856  & 760.49   \\
Meixner-Pollaczek-KAN & 0.9703       & 105{,}862  & 1{,}095.85 \\
Narayana-KAN         & 0.9723       & 105{,}856  & 773.70   \\
Octanacci-KAN        & 0.9688       & 105{,}856  & 770.62   \\
Pado-KAN             & 0.9653       & 105{,}856  & 764.13   \\
Pentanacci-KAN       & 0.9542       & 105{,}856  & 762.46   \\
Tetranacci-KAN       & 0.9675       & 105{,}856  & 767.92   \\
Tribo-KAN            & 0.9569       & 105{,}856  & 751.97   \\
Vieta-Pell-KAN       & 0.9749       & 105{,}856  & 792.53   \\
AskeyWilson-KAN      & 0.9693       & 105{,}871  & 2{,}418.00 \\
\hline
\textbf{SW--KAN (medium)}     & \textbf{0.9753}       & \textbf{105{,}856}  & \textbf{357.0} (28 epochs)    \\
\textbf{SW--KAN (large)}      & \textbf{0.9824}       & \textbf{219{,}907}  & \textbf{221.5} (8 epochs)    \\
\hline
\end{tabular}
\end{table}

\subsection{Fashion-MNIST Classification}
To evaluate the generalization capability and resource efficiency of SW-KAN beyond the MNIST dataset, we conducted a comparative experiment on the Fashion-MNIST benchmark. Fashion-MNIST consists of 60,000 training and 10,000 test grayscale images of 10 fashion categories, with the same dimensionality as MNIST ($28\times28$ pixels). 
Unlike the MNIST experiments, we designed this experiment specifically to assess SW-KAN's performance under resource constrained conditions that are common in real world applications. We applied PCA to reduce the input dimensionality from 784 to 40 components (a 95\% reduction) and subsampled the training set to only 10,000 samples (from the original 60,000), simulating scenarios where computational resources or labeled data are limited. This setup tests the model's data efficiency and robustness to dimensionality reduction.
We compared SW-KAN against the standard B-spline KAN baseline under identical conditions. Both models share the same architecture: a single hidden layer with 48 neurons followed by a 10-class output layer. SW-KAN uses polynomial degree $N=3$, while the B-spline KAN uses grid size 5 and order 3. For a fair comparison with the B-spline baseline, which includes a residual base-activation path, we likewise equip the SW-KAN layer here with a SiLU-based residual path. All models were trained with the Adam optimizer for 30 epochs with learning rate $10^{-2}$ and weight decay $10^{-4}$.

\begin{figure}[H]
    \centering
    \includegraphics[width=1.02\textwidth]{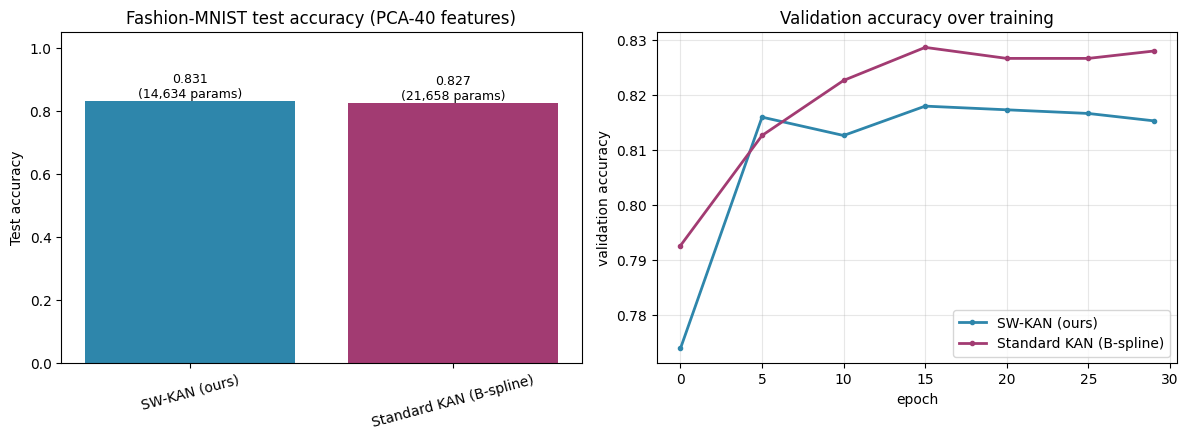}
    \caption{Comparison of SW-KAN and standard B-spline KAN on Fashion-MNIST. Left: test accuracy and parameter counts. Right: validation accuracy during training.}
    \label{fig:fashion_mnist}
\end{figure}

As shown in Figure~\ref{fig:fashion_mnist}, SW-KAN achieves a test accuracy of 83.10\%, outperforming the standard B-spline KAN, which attains 82.65\%. Notably, SW-KAN accomplishes this with significantly fewer parameters (14,634 vs.\ 21,658), representing a 32\% reduction in model size while still delivering superior accuracy. This result is consistent with our MNIST findings and further confirms that Stieltjes-Wigert polynomial basis functions offer a more parameter-efficient representation compared to B-splines.
The validation accuracy curves in Figure~\ref{fig:fashion_mnist} (right) show that both models converge within approximately 15 epochs, with SW-KAN reaching its best validation accuracy slightly earlier than the B-spline KAN. The B-spline KAN achieves its highest validation accuracy of 83.13\% at epoch 9, while SW-KAN reaches 82.60\% at epoch 12. However, SW-KAN maintains competitive performance throughout training and achieves superior final test accuracy despite a slightly lower best validation accuracy, suggesting better generalization to the test set.
Most importantly, these results demonstrate SW-KAN's robustness under three critical resource-constrained conditions:
\begin{itemize}
    \item \textbf{Data efficiency}: With only 10,000 training samples (17\% of the full dataset), SW-KAN maintains competitive performance, indicating strong generalization from limited labeled data.
    \item \textbf{Dimensionality reduction tolerance}: Despite reducing input features from 784 to 40 dimensions (95\% reduction via PCA), SW-KAN preserves its representational power, making it suitable for high-dimensional problems where feature compression is necessary.
    \item \textbf{Parameter efficiency}: With 32\% fewer parameters than the B-spline KAN, SW-KAN achieves higher accuracy, demonstrating superior parameter utilization.
\end{itemize}

\subsection{Function Approximation on 2D Surfaces}

To evaluate SW-KAN's capability in learning continuous multivariate functions, we conducted a function approximation experiment on three distinct 2D surfaces on the domain $[-2, 2]^2$:

\begin{align}
f_1(x,y) &= \sin(x)\cos(y) + 0.5xy \\
f_2(x,y) &= \exp\left(-\frac{x^2 + y^2}{2}\right) \left(1 + 0.3\sin(3x)\cos(2y)\right) \\
f_3(x,y) &= x^2 + y^2 + 0.5\sin(4x) + 0.3\cos(4y)
\end{align}

The SW-KAN architecture consists of a single hidden layer with 16 neurons and polynomial degree $N=4$, resulting in only 389 trainable parameters. Training used 5,000 randomly sampled points for 5,000 epochs with learning rate $10^{-2}$.

\begin{figure}[H]
    \centering
    \hspace*{-0.5cm}
    \includegraphics[width=0.85\textwidth]{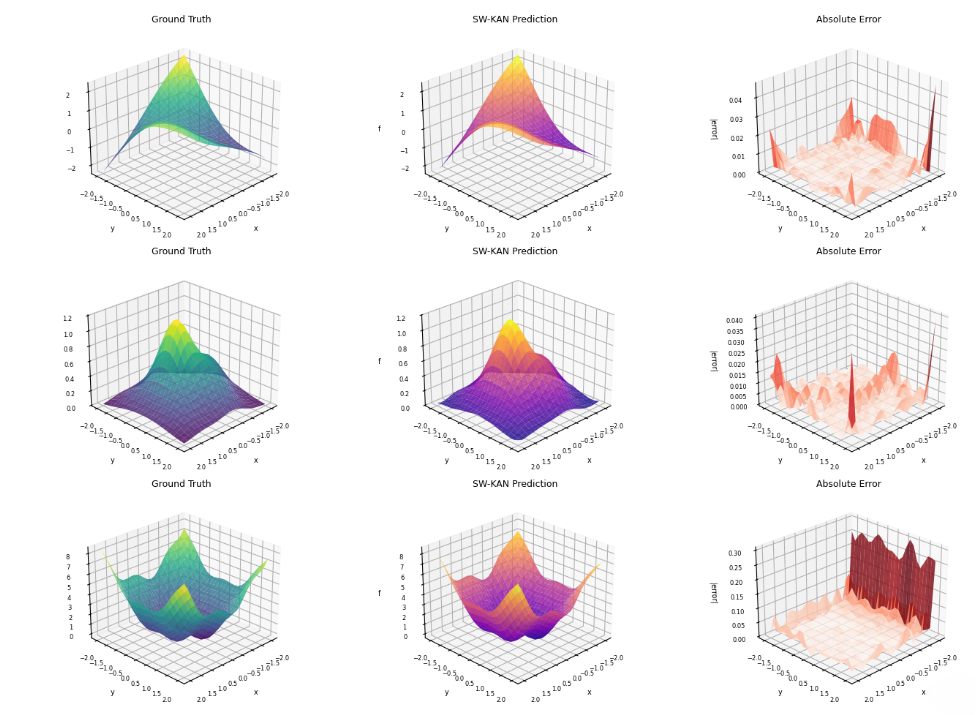}
    \caption{Function approximation results for SW-KAN on three 2D surfaces. Each row corresponds to a different target function: (left) ground truth, (middle) SW-KAN prediction, (right) absolute error.}
    \label{fig:function_approximation}
\end{figure}

Figure~\ref{fig:function_approximation} presents the approximation results. For all three functions, SW-KAN achieves near-perfect reconstruction with minimal errors. The test MSE values are $2.5\times10^{-5}$ for $f_1$, $2.9\times10^{-5}$ for $f_2$, and $3.5\times10^{-3}$ for $f_3$. The error plots confirm that the approximation errors are uniformly distributed across the domain.

These results demonstrate SW-KAN's strong representational power and parameter efficiency, achieving high accuracy across diverse function landscapes with only 389 parameters. The model successfully captures both smooth sinusoidal surfaces and multi-scale oscillatory structures, validating its suitability for scientific computing applications requiring accurate function approximation.

\section{Conclusion and Discussion}
\label{sec:conclusion}
\vspace{0.25cm}

In this paper, we introduced the Stieltjes--Wigert Kolmogorov--Arnold Network (SW--KAN), a novel architecture that integrates $q$-orthogonal polynomials into the KAN framework to address the computational bottlenecks associated with traditional B-spline edge activations. To overcome the structural challenge of mapping unbounded real-valued inputs ($\mathbb{R}$) to the semi-infinite domain $(0,\infty)$ of the Stieltjes--Wigert polynomials, we proposed a smooth exponential-of-tanh transformation. Furthermore, by leveraging a numerically stable, orthonormalized three-term recurrence, we formulated a highly efficient evaluation strategy that computes polynomial expansions in $\mathcal{O}(N)$ operations without relying on computationally expensive special-function calls.

\subsection{Discussion of Experimental Findings}
Our empirical evaluations demonstrate that SW--KAN achieves a superior accuracy-efficiency trade-off compared to existing polynomial-based KANs, and that this advantage generalizes across classification and function-approximation tasks rather than being an artifact of a single benchmark.

On the MNIST handwritten digit classification benchmark, the medium configuration of SW--KAN reached $97.53\%$ test accuracy with $105{,}856$ parameters, already surpassing the best-performing baseline of comparable size, Vieta--Pell-KAN ($97.49\%$ with $105{,}856$ parameters). The large configuration further improved the accuracy to $98.24\%$ using $219{,}907$ parameters, exactly matching Gottlieb-KAN's parameter count yet achieving a higher accuracy ($97.59\%$). Taken together, both the medium and large SW--KAN configurations outperform all 18 established polynomial KAN baselines at their respective parameter budgets.

On Fashion-MNIST, under a deliberately resource-constrained setting (input features reduced from 784 to 40 dimensions via PCA and training limited to $10{,}000$ samples), SW-KAN reached $83.10\%$ test accuracy with $14{,}634$ parameters, outperforming a standard B-spline KAN of the same architecture ($82.65\%$ with $21{,}658$ parameters, a $32\%$ larger model). This indicates that the parameter efficiency observed on MNIST persists, and if anything becomes more pronounced, when training data and feature dimensionality are scarce.

On continuous multivariate function approximation over three qualitatively different 2D surfaces, a SW-KAN model with only $389$ parameters achieved test MSE values as low as $2.5\times10^{-5}$--$2.9\times10^{-5}$ for the smoother targets and $3.5\times10^{-3}$ for the more oscillatory target, confirming that the Stieltjes--Wigert basis is expressive enough for accurate scientific-computing-style regression at a very small parameter budget.

Taken together, the performance delta over baselines cannot be attributed solely to parameter capacity, since the large SW--KAN matches Gottlieb-KAN's parameter count exactly and the Fashion-MNIST SW-KAN model uses fewer parameters than its B-spline counterpart. Instead, this superiority is rooted in the mathematical properties of the basis functions and their interaction with gradient-based optimization. Classical orthogonal polynomials are typically defined on strictly bounded intervals (e.g., $[-1,1]$) or discrete domains. When standard normalizations (like $\tanh$ or sigmoid) are used to map unbounded network activations into these fixed domains, the inputs frequently saturate at the boundaries, leading to severe gradient attenuation (vanishing gradients) for heavy-tailed distributions. Conversely, the Stieltjes--Wigert polynomials are naturally supported on $(0, \infty)$. By employing the smooth exponential-of-tanh mapping, inputs are projected into a safe, strictly positive sub-interval that completely avoids zero and extreme asymptotic tails.

Furthermore, unlike rigid classical families, the learnable $q$ parameter in SW--KAN acts as a dynamic frequency modulator during backpropagation. This geometric flexibility, combined with the log-normal weight structure, yields a smoother and more well-conditioned loss landscape, which is consistent with the model's competitive results across image classification and continuous function approximation despite comparable or smaller parameter budgets than its baselines.

\subsection{Limitations and Future Work}
Despite the promising results, this study exhibits certain limitations that provide avenues for future research. First, while our empirical validation now spans MNIST, Fashion-MNIST under resource-constrained conditions, and continuous function approximation on 2D surfaces, it remains focused on relatively small-scale, low-to-moderate-dimensional tasks. Evaluating SW--KAN on larger, higher-dimensional problems, such as ImageNet-scale image classification, dense time-series modeling, and physics-informed neural network (PINN) applications, is necessary to fully assess its scalability.

Second, the domain mapping parameters (e.g., the scaling factors in the exponential-of-tanh function) were treated as fixed hyperparameters. Future iterations of SW--KAN could benefit from making these domain-mapping parameters fully learnable, enabling dynamic, layer-specific distribution matching. A related direction is to relax the current layer-wise sharing of $q^{(\ell)}$ (Section~\ref{subsec:sw-learnable-params}) toward finer-grained, e.g.\ edge-wise or group-wise, $q$ parameterizations, and to characterize the resulting accuracy-versus-parameter-count trade-off relative to the layer-shared design studied here. 

Finally, a distinguishing mathematical feature of the Stieltjes--Wigert polynomials is their indeterminate moment problem, meaning multiple distinct weight measures yield the same orthogonality relation. The theoretical implications of this indeterminacy on the loss landscape, implicit regularization, and generalization bounds of deep neural networks remain an open question. Future theoretical work will focus on understanding how this indeterminacy influences the optimization dynamics of $q$-orthogonal KANs.

\section*{Data and Code Availability}
The source code, configuration files, and pretrained models required to reproduce the experimental results presented in this paper are publicly available at \href{https://github.com/amirhoseinazarpour/SW-KAN}{https://github.com/amirhoseinazarpour/SW-KAN}.

\bibliographystyle{IEEEtran}
\bibliography{sample}

\end{document}